\documentclass[journal]{IEEEtran}
\usepackage[font=small,skip=1pt]{caption}
\IEEEoverridecommandlockouts
\usepackage{textcomp}
\usepackage{gensymb}
\usepackage{latexsym}
\usepackage[T1]{fontenc}
\usepackage{multirow}
\usepackage{graphicx}
\usepackage{varioref}
\usepackage{array}
\usepackage{enumitem}
\usepackage{soul,color}
\usepackage{longtable}
\newcommand{\cmark}{\ding{51}}%
\newcommand{\xmark}{\ding{55}}%
\usepackage{algorithmic}
\usepackage[ruled, vlined]{algorithm2e}
\usepackage{pifont}
\usepackage{booktabs}

\ifCLASSOPTIONcompsoc
  \usepackage[caption=false,font=normalsize,labelfont=sf,textfont=sf]{subfig}
\else
  \usepackage[caption=false,font=footnotesize]{subfig}
\fi

\usepackage{color}
\definecolor{light}{rgb}{0.5, 0.5, 0.5}

\usepackage{graphicx}
\medskip
\usepackage{amssymb}
\usepackage{amsmath,epsfig,amssymb}
\usepackage{amsthm}
\usepackage{xcolor}
\usepackage[active]{srcltx} 
\usepackage{epstopdf} 

\usepackage{bm}
\usepackage{amsthm}
\usepackage{enumitem}
\DeclareRobustCommand*{\IEEEauthorrefmark}[1]{%
  \raisebox{0pt}[0pt][0pt]{\textsuperscript{\footnotesize\ensuremath{#1}}}}

\usepackage{cite}
\usepackage{amsmath,amssymb,amsfonts}
\usepackage{algorithmic}
\usepackage{graphicx}
\usepackage{textcomp}
\usepackage{xcolor}
\def\BibTeX{{\rm B\kern-.05em{\sc i\kern-.025em b}\kern-.08em
    T\kern-.1667em\lower.7ex\hbox{E}\kern-.125emX}}
\usepackage{lineno,hyperref}
\usepackage[utf8]{inputenc}
\usepackage[english]{babel}
\usepackage{mathdots}

\graphicspath{{Pictures/}{../jpeg/}} 
\usepackage{float}
\usepackage{amsmath}

\begin{document}

\title{MpoxSLDNet: A Novel CNN Model for Detecting Monkeypox Lesions and Performance Comparison with Pre-trained Models}

\author{\IEEEauthorblockN{Fatema Jannat Dihan\IEEEauthorrefmark{1}, Saydul Akbar Murad\IEEEauthorrefmark{2}}

\thanks{\IEEEauthorrefmark{1} Noakhali Science and Technology University, Sonapur, Noakhali, 3814, Bangladesh, \IEEEauthorrefmark{2} School of Computing Sciences \& Computer Engineering, University of Southern Mississippi, Hattiesburg, MS, USA}
}  




\maketitle
\begin{abstract}
Monkeypox virus (MPXV) is a zoonotic virus that poses a significant threat to public health, particularly in remote parts of Central and West Africa. Early detection of monkeypox lesions is crucial for effective treatment. However, due to its similarity with other skin diseases, monkeypox lesion detection is a challenging task. To detect monkeypox, many researchers used various deep-learning models such as MobileNetv2, VGG16, ResNet50, InceptionV3, DenseNet121, EfficientNetB3, MobileNetV2, and Xception. However, these models often require high storage space due to their large size. This study aims to improve the existing challenges by introducing a CNN model named  MpoxSLDNet (Monkeypox Skin Lesion Detector Network) to facilitate early detection and categorization of Monkeypox lesions and Non-Monkeypox lesions in digital images. Our model represents a significant advancement in the field of monkeypox lesion detection by offering superior performance metrics, including precision, recall, F1-score, accuracy, and AUC, compared to traditional pre-trained models such as VGG16, ResNet50, and DenseNet121. The key novelty of our approach lies in MpoxSLDNet's ability to achieve high detection accuracy while requiring significantly less storage space than existing models. By addressing the challenge of high storage requirements, MpoxSLDNet presents a practical solution for early detection and categorization of monkeypox lesions in resource-constrained healthcare settings. In this study, we have used "Monkeypox Skin Lesion Dataset" comprising 1428 skin images of monkeypox lesions and 1764 skin images of Non-Monkeypox lesions. Dataset's limitations could potentially impact the model's ability to generalize to unseen cases. However, the MpoxSLDNet model achieved a validation accuracy of 94.56\%, compared to 86.25\%, 84.38\%, and 67.19\% for VGG16, DenseNet121, and ResNet50, respectively. In addition, an average precision, recall, F1 score of 0.94 respectively, and AUC of 0.94 was recorded for our proposed model.
\end{abstract} 

\section{Introduction}

\begin{figure}[t]
    \centering
    \includegraphics[width=\linewidth]{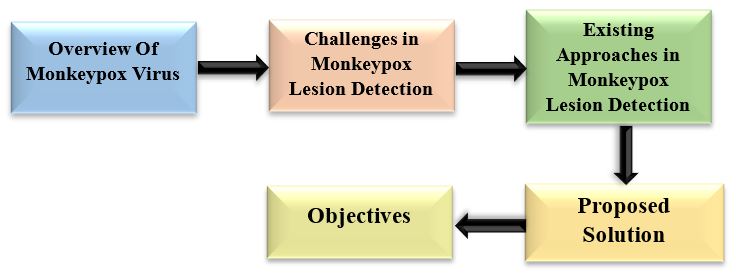}
    \caption{Navigating Monkeypox Lesion Detection: An Introduction Overview.}
    \vspace{-10pt}
    \label{fig1}
\end{figure}

 \textit{Figure \ref{fig1}} provides an introductory overview of the efforts to enhance monkeypox lesion detection. It begins by elucidating the characteristics of the Monkeypox virus including its origins, classification, symptoms, and modes of transmission. Monkeypox virus (MPXV) is a zoonotic virus that is predominantly found in distant Central and West African regions \cite{}. It was first identified in a monkey in 1959, when outbreaks of a disease resembling pox were observed in monkeys retained for research purposes, in a laboratory in Copenhagen, at a Danish research institute, and thus the name Monkeypox virus \cite{R2, R3}. It is a type of DNA virus classified within the orthopoxvirus genus and is a member of the family of Poxviridae, is closely
linked to the Variola virus, which is the infectious illness that
causes monkeypox \cite{R4}. In 1970, the first recorded instance of monkeypox cases in humans was documented in the Democratic Republic of the Congo, within an intensified campaign to wipe out smallpox. The monkeypox virus falls into two groups: Congo Basin (Central African) and West African. According to a previous report, the Congo Basin clade tends to produce severe disease more often, with case fatality ratios (CFR) of up to 10\%. As of right now, the Democratic Republic of Congo reports a 3\% Case Fatality Ratio among suspected cases. In the past, the West African clade has been attributed to a lower overall CFR of roughly 1\% in an African population who are normally younger. An average incubation period is 12 days for MPXV, yet it might vary from 5 to 21 days. \cite{R5}. Monkeypox virus signs involve fever, headache, tenderness in the muscles, backaches, swelling of the lymph nodes, chills, as well as exhaustion \cite{R6,R7, new2}. The monkeypox virus circulates from a human to another human being thru droplets from the nose or throat, interaction with affected bodily fluids or contaminated surfaces, infectious rash, and sexual contact. This virus can also spread through the handling of infected animals \cite{R8}. 

There is unfortunately no medication or vaccination specifically against the monkeypox virus. Although supportive care can be given to combat symptoms and undesirable side effects. Antiviral drugs can be used that are used for treating other orthopoxvirus infections \cite{R9}. A skin lesion test utilizing the technique of electron microscopy or polymerase chain reaction, also known as a PCR test is time-consuming and takes a long time to detect pox \cite{R10}. The insufficient availability of PCR test kits, as well as the strict requirements for laboratory settings in research facilities, will delay the exact identification of suspects \cite{R11}. 

The detection of monkeypox lesions can be challenging due to the similarity that exists between monkeypox lesions and another pox \cite{R12, R13}, which can lead to misdiagnosis. In addition, personnel in healthcare might have been unfamiliar with a clinical presentation of the lesions of monkeypox \cite{R14}, which can further complicate accurate diagnosis. Moreover, physical examinations to diagnose monkeypox can increase the risk of transmission to healthcare professionals \cite{R15} and other patients. According to \cite{R16}, isolation is recommended for a patient with a suspected monkeypox infection. 

To address all these challenges, there is a need for the development of a reliable and accurate diagnostic tool that can quickly and non-invasively detect and classify monkeypox lesions from other poxvirus infections which also can significantly reduce the risk of transmission to healthcare professionals and other patients. Neural networks can help in this situation. With advances in the development involving AI models, many models based on deep learning have been utilized in the identification of various diseases \cite{R17, R18, R19, new1}. So, in our current research, we have developed a model named “MpoxSLDNet” based on convolutional neural networks (CNNs) to automatically detect monkeypox lesions from digitized pictures of human skin lesions.

From the existing research work regarding monkeypox lesion detection, we can observe, in \cite{R1}, authors used publicly available datasets. They presented an Android mobile application in which low modified deep learning model MobileNetv2 is used to classify the monkeypox lesion and non-monkeypox lesions with 91.11\% accuracy. In \cite{R11}, authors developed a monkeypox skin lesion dataset, and they employed deep learning networks including VGG16, ResNet50, InceptionV3, and Ensemble. In their study, among the evaluated models, ResNet50 demonstrated the highest overall accuracy, achieving an accuracy of 82.96\%. Also VGG16, InceptionV3,Ensemble obtained accuracy of 81.48\%, 74.07\%,79.26\% respectively. Here, the authors presented a web app that used the best-performing model. Apart from this, Authors \cite{R21} used publicly available datasets and employed five deep learning models, including DenseNet121, VGG19, MobileNetV2, EfficientNetB3, and Xception combined with the CBAM and Dense layers for categorization. Here, VGG19, DenseNet121, EfficientNetB3, MobileNetV2, and Xception achieved accuracy 71.86\%, 78.27\%, 81.43\%, 74.07\%, 83.89\% respectively. Despite the significant efforts in the existing research work regarding monkeypox lesion detection, our model achieves superior results on the basis of accuracy, precision, recall, F1-score, and AUC compared to the state-of-the-art models. While the existing thesis work on monkeypox lesion detection has employed several deep-learning models, these models require significant storage space due to their large number of parameters and also require more computational time. In contrast, our proposed CNN model is designed to be both accurate and efficient, with fewer parameters that require a smaller amount of storage as well as computational time. This makes our approach more practical for use in resource-constrained settings, such as remote areas where monkeypox outbreaks are more likely to occur.

Our proposed MpoxSLDNet model is a Sequential neural network containing max pooling, convolutional layers, batch normalization, dropout, as well as fully connected layers. It takes input images of size 224x224x3 and learns to classify them as either having monkeypox lesions or not. The feature extraction process in our model involves using a series of convolutional layers of adding depth  in addition to reducing spatial dimensions, commencing with max pooling layers to reduce the dimensionality of the feature maps. The extracted features are then flattened and passed through various fully connected layers with batch normalization and dropout regularization to stay away from overfitting. The final layer of output uses a sigmoid activation function to acquire a binary classification result. The contribution of this study to the field of monkeypox lesion detection is listed below:
\begin{itemize}
    \item Developed 'MpoxSLDNet', a convolutional neural network model, for automated identification of monkeypox lesions in digitized human skin lesion imagery, utilizing advanced image processing and deep learning techniques.
\end{itemize}
\begin{itemize}
    \item Validated 'MpoxSLDNet's effectiveness in precisely distinguishing monkeypox lesions from other poxvirus infections, outperforming existing pre-trained models in accuracy through rigorous dataset comparisons and performance metrics analysis.
\end{itemize}
\begin{itemize}
    \item Introduced a contactless, non-invasive diagnostic approach, significantly reducing transmission risks to healthcare professionals and patients, utilizing advanced sensor technologies and AI-driven analysis.
\end{itemize}

The rest of this study is organized as follows: section 2 describes and discusses the literature review. Section 3 provides details about the methodology in which several methodological steps were undertaken, including data collection, preprocessing, the architecture of the MpoxSLDNet model, and the evaluation method. Section 4 evaluates the results. This section provides a comprehensive analysis of the outcomes obtained from our suggested model and compares them with pre-trained models, including DenseNet121, ResNet50, and VGG16, highlighting its effectiveness in detecting monkeypox lesions. Finally, this paper is concluded in Section 5.

\section{Literature Review}

Monkeypox is a rare and serious viral disease that presents a significant public health risk. The timely detection of monkeypox lesions is essential for effective treatment as well as outbreak prevention. Deep learning-based models have gained prominence in recent years due to their potential use in automating the detection of monkeypox lesions. A recent study conducted by \cite{R4} collected monkeypox images from various sources and developed a dataset of monkeypox images that are publicly available. This study established a modified VGG16 model to classify monkeypox patients and non-monkeypox patients. Post-image analysis explanations using a technique called Local Interpretable Model-Agnostic Explanations (LIME) were also used in order to verify outcomes. The modified VGG16 model was evaluated for two distinct studies, including study one (without augmentation) and study two (with augmentation). The achieved accuracies for studies one and two were 97 ± 1.8\% and 88 ± 0.8\%, respectively. However, this study only considered the modified VGG16 model. Additionally, Another study conducted by \cite{R22}, proposed a model named MonkeypoxHybridNet based on CNN, which is comprised of VGG19, ResNet50, and InceptionV3. The performance of the model was assessed by evaluating its precision, F1 score, and accuracy against three pre-trained models, including VGG19, ResNet50, and InceptionV3. The achieved accuracy for VGG19, ResNet50, InceptionV3, and MonkeypoxHybridNet were 0.595, 0.705, 0.80, and 0.842, respectively. The proposed model showed significant results, but it has some limitations. The authors only presented accuracy, precision, F1 score, and confusion matrix but did not show the Recall and AUC score for the proposed MonkeypoxHybridNet model. In a subsequent study by \cite{R14}, a variety of pre-trained models were used to distinguish between chickenpox and monkeypox lesions, including Inception V3, ResNet50, VGG16, VGG19, and AlexNet. They also implemented a CNN model. The chickenpox dataset used in the study was publicly available, while the monkeypox dataset was obtained through web scraping.VGG16 and VGG19 accuracy was also 80\%, while ResNet50, AlexNet, and Inception V3 each provided accuracy results of 82.00\%, 98.00\%, and 89.00\%, respectively. The recorded accuracy of the proposed CNN model was 99\%. However, the AUC score, which is an important measure of classifier performance, was not calculated in this study. Apart from this, another study by \cite{R2}  compared the performance of 13 pre-trained models to detect the monkeypox lesion. Universal custom layers had been used for all models, and the performance was analyzed based on four metrics: accuracy, precision, recall, and F1-score. The study identified DL models that exhibited the highest performance and ensembled them via majority voting on their probabilistic outputs to enhance the overall performance. The experiments utilized a dataset that is publicly accessible and the proposed ensemble method provided accuracy, precision, recall, and F1-score of 87.13\%, 85.44\%, 85.47\%, and 85.40\%, respectively. However, this study did not propose any lightweight CNN models, and the AUC score was not considered. In a subsequent study conducted by \cite{R11}, A dataset called the Monkeypox Skin Lesion Dataset (MSLD) was created, which includes images of skin lesions associated with monkeypox, chickenpox, and measles. Three widely used pre-trained models, including InceptionV3, ResNet50, and VGG16, were utilized to identify monkeypox and non-monkeypox patients. Additionally, an ensemble of three models (InceptionV3, ResNet50, and VGG16) was employed in this study. The achieved accuracies were 81.48\% (±6.87\%), 82.96\% (±4.57\%), 74.07\% (±3.78\%), and 79.26\% (±1.05\%) for VGG16, ResNet50, InceptionV3, and the ensemble method, respectively. Furthermore, a web application was employed as an online monkeypox screening tool. It is known that pre-trained models require more storage space and computational time. In this work, only pre-trained models were utilized, and no lightweight CNN models were established. As for performance evaluation metrics, the AUC score was not considered. Furthermore, a study conducted by \cite{R24}, used five pre-trained models including VGG16, VGG19, MobileNetV2, ResNet50, and EfficientNetB3. The performance of the model is assessed by measuring its accuracy, recall, precision, and F1 score. Their experimental result demonstrates that among the five pre-trained models MobileNetV2 gives the best classification performance. Achieved accuracy, precision, recall, and F1-score of MobileNetV2 are 98.16\%, 0.99, 0.96, and 0.98. Their objective was to select the best-performing model to detect and classify monkeypox. In this study, the authors don't consider the AUC score and don't show the multiple training and testing accuracy. They also don't discuss losses during training. Another study employed by \cite{R25}, a newly curated image dataset, referred to as the "Monkeypox skin images dataset," was developed, encompassing four classes: monkeypox, chickenpox, measles, and normal images. In this study, five machine learning classifiers, including Logistic Regression, Random Forest, Support Vector Machine, K-Nearest Neighbor, as well as Extreme Gradient Boosting were used for classification. For feature extraction, VGG16, ResNet50, MobileNetV1, Inception V3, Xception. Also, they presented a novel CNN model called MonkeyNet, which is a modified version of DenseNet-201, and evaluated its performance. The performance of the model was assessed using various metrics including accuracy, precision, recall, F1-score, and AUC. Resulted in accuracy for the original dataset and augmented dataset of the proposed model are 91.91\% and  98.91\%, consecutively. Another study conducted by \cite{R21} utilized five deep learning models including VGG19, Xception, DenseNet121, MobileNetV2, and EfficientNetB3.To utilize a more relevant feature map, they integrate the deep learning model with the Convolutional Block Attention Module. This module encompasses mechanisms such as channel attention and spatial attention-based mechanism. Then dense layer, flattened layer, and classification layer are added to each model to classify monkeypox and others. The model's performance is assessed using accuracy, precision, recall, and F1-score as evaluation metrics. Among the various combinations, Xception-CBAM-Dense yielded the highest performance with a validation accuracy of 83.89\%, precision of 90.70\%, recall of 89.10\%, and F1-score of 90.11\%. Authors \cite{R32} used deep learning techniques, specifically deep neural networks, for the detection of monkeypox virus using skin lesion images. They tested the dataset on five pre-trained deep neural networks: GoogLeNet, Places365-GoogLeNet, SqueezeNet, AlexNet, and ResNet-18. Transfer learning models such as Xception, DenseNet-169, and MobileNetV2 were also employed in other studies related to monkeypox diagnosis. Data augmentation techniques were applied to increase the size of the dataset and improve the performance of the deep neural networks. The deep learning models achieved high accuracy in detecting monkeypox virus from skin lesion images, with ResNet-18 obtaining the highest accuracy of 99.49\%. Other performance metrics such as precision, recall, f1-score, and AUC were also considered, and the modified models obtained validation accuracies above 95\%. Authors \cite{R33} employed deep convolutional neural networks of the residual network family, including ResNet-18, ResNet-50, and ResNet-101, to solve the vanishing gradient problem and achieve high performance in image classification. They also used SqueezeNet as a baseline model. The researchers achieved an average accuracy of 91.19\% and an F1-score of 92.55\% for the Monkeypox class using deep learning models. The researchers also used explainable artificial intelligence (XAI) techniques, such as LIME (Local Interpretable Model-Agnostic Explanations), to visualize the predictions made by the convolutional neural network. Authors \cite{R33} integrate deep neural networks with federated learning (FL) to address the challenges of medical data categorization. Also,they used a cycle-consistent generative adversarial network (GAN) to augment data samples for training. And utilize deep learning models such as MobileNetV2, Vision Transformer (ViT), and ResNet50 for the classification task. Authors \cite{R35} developed a hybrid artificial intelligence system using deep learning models (CSPDarkNet, InceptionV4, MnasNet, MobileNetV3, RepVGG, SE-ResNet, and Xception) and a long short-term memory (LSTM) model for monkeypox detection. The classification results were analyzed using metrics such as test accuracy and Cohen's kappa score.The proposed hybrid artificial intelligence system achieved a test accuracy of 87\% and a Cohen's kappa score of 0.8222 for monkeypox detection. \cite{R36} presented a model, PoxNet22, by fine-tuning the InceptionV3 to classify monkeypox from other poxes. They used different preprocessing techniques to improve the image quality and used augmentation to reduce the possibility of overfitting. They got very high precession, recall and accuracy. \cite{R37} used transfer learning to assess the performance of different DL mehtods namely, VGG16, InceptionResNetV2, ResNet50, ResNet101, MobileNetV2, VGG19, and Vision Transformer (ViT) for diagnosing monkeypox. Among their models, modified VGG16 and MobileNetV2 outperformed others. For proving model's transparency, they used LIME with their model. \cite{R38} presented a model for detecting monkeypox by using attention-based MobileNetV2. They used diversified performance parameters to asses the superiority of the proposed model. They also used Grad-CAM and LIME for ensuring transparency and interpretability of their model.    

\textit{Table \ref{Table1}} presents a  comprehensive comparison of various parameters employed in previous studies and our study focused on monkeypox lesion detection. Each row corresponds to a specific study, while the columns represent different evaluation metrics that include accuracy, precision, recall, F1-score, area under the curve (AUC), etc. This analysis offers valuable insights into the existing literature and highlights the unique contribution of our study by incorporating all key evaluation metrics for a comprehensive evaluation of monkeypox lesion detection. Our study aims to contribute to the existing literature on monkeypox lesion detection by providing a comprehensive comparison of key evaluation metrics, including multiple training accuracy,multiple testing accuracy, Training loss, Testing Loss, precision, recall, F1-score, and area under the curve (AUC). As demonstrated in \textit{Table \ref{Table1}}, our analysis revealed that several performance metrics were not consistently reported across existing studies. This observation underscores the need for a more standardized and comprehensive approach to evaluating monkeypox lesion detection models. By systematically comparing the performance of various pre-trained models and proposing a novel lightweight CNN model, MpoxSLDNet, our study addresses the limitations identified in previous research. Specifically, we have incorporated all key evaluation metrics, including multiple training accuracy,multiple testing accuracy, Training loss, Testing Loss, precision, recall, F1-score, and area under the curve (AUC), which were often overlooked in previous studies. This holistic evaluation approach provides a more robust assessment of model performance and facilitates a deeper understanding of the strengths and weaknesses of different monkeypox lesion detection methodologies.

\begin{table*}[htb]
    \caption{Comparative analysis of parameters used in existing studies with our study.}
    \resizebox{\linewidth}{!}{
\label{Table1}
\begin{scriptsize}
\centering
\begin{tabular}[c]{|  p{2.8cm} | p{1.2cm}  | p{1.2cm}  | p{.8cm}  | p{.8cm}  | p{.8cm}  | p{.8cm}  | p{.8cm}  | p{.9cm}  | p{1.2cm}  |}
\hline   

\hline
\textit{Ref.} & \textit{Multiple training accuracy} & \textit{Multiple Testing Accuracy} & \textit{Training Loss} & \textit{Testing Loss} & \textit{Precision} & \textit{Recall} & \textit{F1-Score } & \textit{Accuracy} & \textit{AUC-ROC Curve} \\ 
\hline

\hline
\cite{R4}  & \xmark &	\xmark &	\xmark &	\xmark &	\cmark &	\cmark & \cmark &	\cmark &	\cmark\\
\hline
\cite{R22} & \xmark &	\xmark &	\xmark &	\xmark &	\cmark &	\xmark &	\cmark &	\cmark &	\xmark \\
\hline
\cite{R14} & \xmark &	\xmark &	\xmark &	\cmark &	\cmark &	\cmark &	\cmark &	\cmark &	\xmark \\
\hline
\cite{R2} & \xmark &	\xmark &	\cmark &	\cmark &	\cmark &	\cmark &	\cmark &	\cmark &	\xmark \\
\hline
\cite{R11} & \xmark &	\xmark &	\xmark &	\xmark &	\cmark &	\cmark &	\cmark &	\cmark &	\xmark \\
\hline
\cite{R24} & \xmark &	\xmark &	\xmark &	\xmark &	\cmark &	\cmark &	\cmark &	\cmark &	\xmark \\
\hline
\cite{R21} & \xmark &	\xmark &	\xmark &	\xmark &	\cmark &	\cmark &	\cmark &	\cmark &	\xmark \\
\hline
\cite{R33} & \xmark &	\xmark &	\cmark &	\cmark &	\cmark &	\cmark &	\cmark &	\cmark &	\xmark \\
\hline
\cite{R32} & \xmark &	\xmark &	\xmark &	\xmark &	\cmark &	\cmark &	\cmark &	\cmark &	\xmark \\
\hline
Proposed Method & \cmark &	\cmark &	\cmark &	\cmark &	\cmark &	\cmark &	\cmark &	\cmark &	\cmark \\
\hline

\end{tabular}
\end{scriptsize} }
\end{table*}

\section{Methodology}
\begin{figure}[t]
    \centering
    \includegraphics[width=1\linewidth]{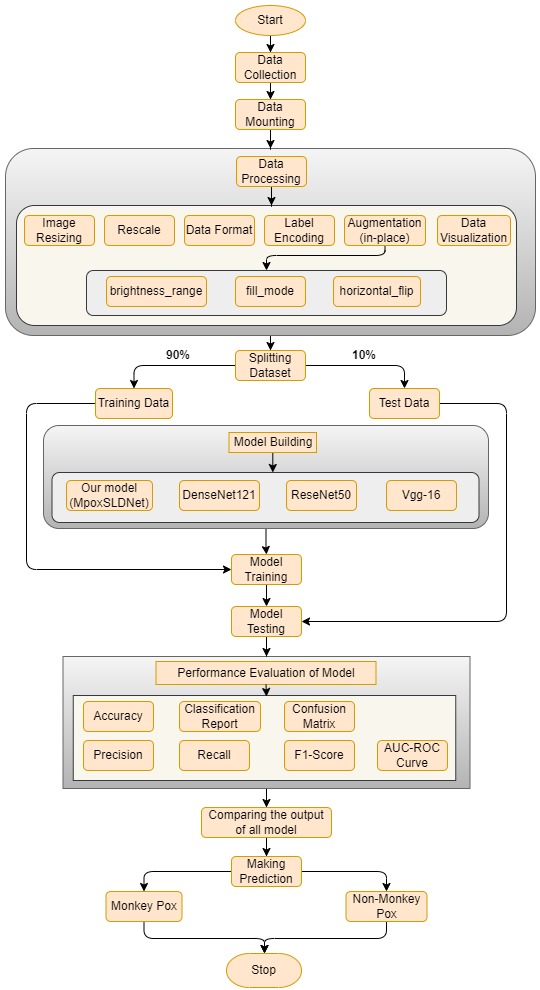}
    \caption{Overall Workflow diagram.}
    \vspace{-10pt}
    \label{fig2}
\end{figure}

The methodology section of this study outlines the process of data collection, data mounting, data pre-processing, the proposed MpoxSLDNet model's structure, and the evaluation metrics used to compare the performance of MpoxSLDNet with three pre-trained models, including VGG16, ResNet50, and DenseNet121. Firstly, to achieve the objectives that are set out, the dataset of monkeypox skin lesion images was collected and pre-processed to ensure consistency and quality. Then the suggested CNN model MpoxSLDNet was built and three pre-trained models including VGG16, ResNet50, and DenseNet121 were utilized for transfer learning. Finally, The MpoxSLDNet model's performance is assessed by comparing it to the pre-trained models using standard evaluation metrics including accuracy, precision, recall, and F1-Score, AUC-ROC curve. The overall workflow of this study is shown in \textit{Figure \ref{fig2}}.

\subsection{Data Collection}

\begin{figure}[t]
    \centering
    \includegraphics[width=\linewidth]{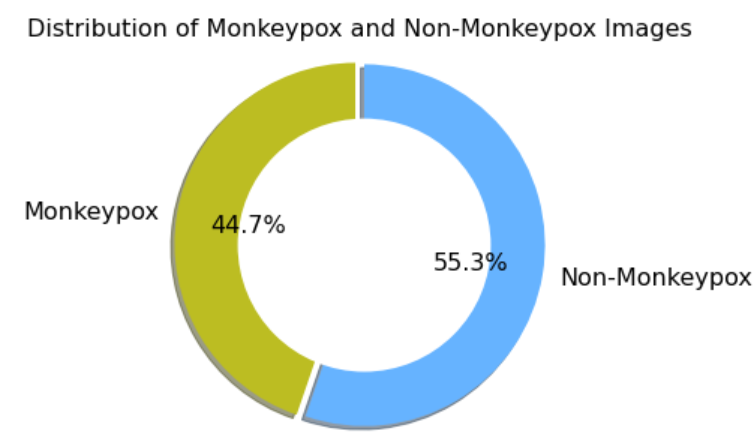}
    \caption{Distribution of Monkeypox lesion and Non-Monkeypox lesion images in the dataset.}
    \vspace{-10pt}
    \label{fig3}
\end{figure}

\begin{figure}[t]
    \centering
    \includegraphics[width=\linewidth]{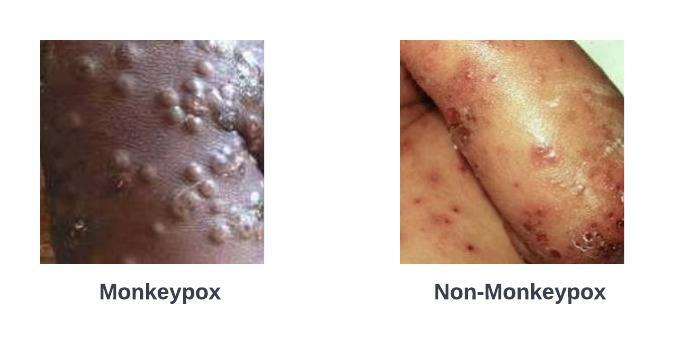}
    \caption{Sample set of images from dataset.}
    \vspace{-10pt}
    \label{fig4}
\end{figure}

In this study, data are collected from a publicly accessible source known as the "Monkeypox Skin Lesion Dataset" \cite{R26}. The dataset includes two classes: Monkeypox and Non-Monkeypox, which are used for performing binary classification. The Monkeypox class contains 1428 skin images of monkeypox lesions. On the other hand, Non-Monkeypox class contains 1764 skin images of Non-Monkeypox lesions. The detailed data description is shown in \textit{Table \ref{Table2}}. \textit{Figure \ref{fig3}} illustrates a visual representation of the dataset's image distribution, highlighting the proportion of monkeypox lesion images compared to non-monkeypox lesion images. This information is crucial for understanding the composition of the dataset and the relative representation of different lesion types, thus providing valuable insights into the data used for training and evaluation in our study. \textit{Figure \ref{fig4}} shows the Sample set of images of Monkeypox lesion and Non-Monkeypox lesion from the dataset.

\footnote{https://www.kaggle.com/datasets/nafin59/monkeypox-skin-lesion-dataset}

\begin{table}[htb]
    \caption{Dataset Description.}
    \resizebox{\linewidth}{!}{
\label{Table2}
\begin{scriptsize}
\centering
\begin{tabular}[c]{|  p{1.2cm} | p{1.8cm}  | p{1.2cm} |}
\hline

\hline
\textit{Class} & \textit{Description} & \textit{Source of data} \\ 
\hline

\hline
Monkeypox & In this dataset, the monkeypox class contains 1428 skin images of monkeypox lesions. & Kaggle Datasets
\cite{R26}	 \\
\hline
Non-Monkeypox & Non-Monkeypox class contains 1764  Skin images of chickenpox lesions, measles lesions. & Kaggle Datasets
\cite{R26}	 \\
\hline
\end{tabular}
\end{scriptsize} }
\end{table}

\subsection{Data Pre-processing}
This step is crucial for the deep learning model as it ensures that the input data is appropriately formatted for training the model, leading to enhanced model accuracy. In this step, collected data are passed through six pre-processing operations before it is fed into the model. These pre-processing operations include resizing images to uniform sizes, rescaling the pixel value, setting data\_format, label encoding, in-place augmentation, and data  \cite{new3}. All pre-processing operations are described here.

\subsubsection{Resizing Image}
Image resizing is the process of resampling or downsampling the image. Images in the dataset should have the same size as the CNN model. So, these stages are used to assure that all the images have the same size. This stage is also used to reduce the computational burden of training the model. This is accomplished with the aid of Python/TensorFlow.keras.preprocessing module. In this study, image dimensions are set to 224×224 dimension. The reason for selecting image size 224×224 is that this size allows better performance, more accurate training, and faster convergence of model \cite{R28}. 

\subsubsection{Rescale: }
Rescaling is the process of adjusting or normalizing the image data to a specific range of 0 to 1. ImageDatagenerator class is used to transform the pixel values of images from the original range of 0 to 255 into the desired range of 0 to 1, which is more suitable for neural network models. Different image has different pixel value. If rescale operation is not applied, the model only encounters certain images which leads to overfitting. This operation ensures that each image is given equal importance and weight during the model’s training process. This operation can be accomplished by specifying the rescale argument which involves multiplying each pixel by a specific ratio to achieve the desired scaling. In this scenario, the ratio used for rescaling the pixel values is 1/255, which is approximately equal to 0.0039 \cite{R24}.

\subsubsection{Data\_format: }
In this study, data\_format is set to "channels\_last" which means that the last dimensions of the input tensor are channels. In this "channels\_last" format image is depicted as a 3D tensor that corresponds to (height, width, and channels). In deep learning frameworks like TensorFlow and Keras, this format is the default value in Keras. During training and evaluating the model, this format ensures compatibility, efficient computation, and data manipulation \cite{R29}.

\subsubsection{Label Encoding:  }
Label encoding is a process of transforming categorical labels into numerical representations, enabling deep-learning predictive models to comprehend them effectively. In this study, the dict and zip method is used to create a mapping between the two classes ‘Monkeypox’ and ‘Non\_Monkeypox’ and their corresponding numerical values 0 and 1. By using the dict method, we have created a dictionary in which the keys are the numerical values and the values are corresponding categorical labels. The zip method then makes a list of rows by combining the numerical values with the categorical labels. With the resulting dictionary, the categorical labels can be transformed into numerical values, allowing the model to utilize these numerical representations for making predictions. \cite{R27}.

\subsubsection{Data Augmentation (in-place):}
\begin{figure}[t]
    \centering
    \includegraphics[width=\linewidth]{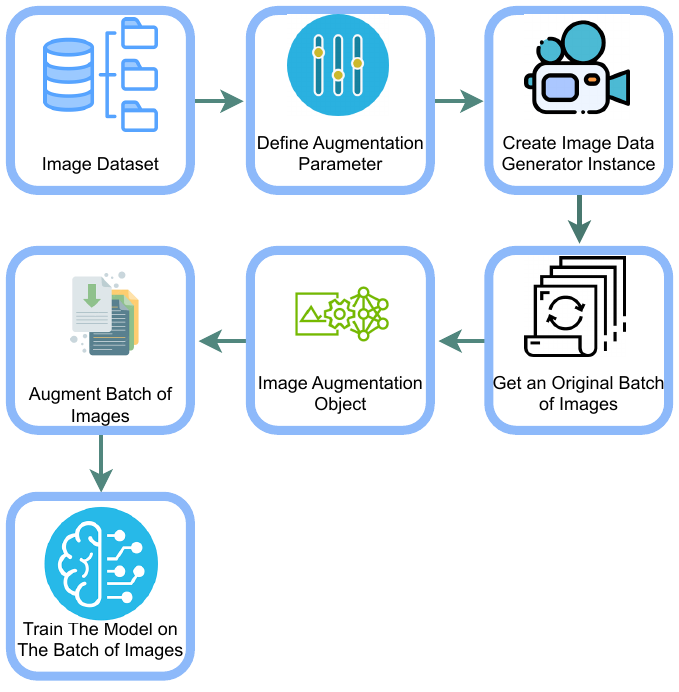}
    \caption{Image Augmentation (in-place ) process.}
    \vspace{-10pt}
    \label{fig5}
\end{figure}

In this pre-processing operation, random image transformations performed by in-place data augmentation (on-the-fly) have been applied, which means that the dataset size does not increase. In this type of data augmentation, at the time of training, our network observes unique variations of our data during each epoch and captures different patterns and features with each iteration. To transform the image, ImageDatagenerator, a Keras image prepossessing library, is used. Image transformation operations available in the ImageDatagenerator function include brightness\_range, zoom range, horizontal flip, etc. \textit{Table \ref{Table3}} depicts the augmentation technique used in this study. \textit{Figure \ref{fig5}}  shows the in-place augmentation process \cite{R30}. The process of applying in-place data augmentation is \cite{R30}:

\textbf{Step-1:} At first, augmentation parameters including zoom\_range, brightness\_range, fill\_mode, and horizontal\_flip are defined.

\textbf{Step-2:} An instance of an ImageDataGenerator class, specifying the augmentation parameters is also created.

\textbf{Step-3:} An original batch of images is passed to the ImageDataGenerator class.

\textbf{Step-4:} Then, the batches of augmented data are generated and returned to the calling method.

\textbf{Step-5:} After that, the model is trained on batches of images.

\begin{table}[]
\caption{Augmentation technique used in this study.}
\label{Table3}
\centering
\begin{tabular}{|c|c|}
\hline
\textbf{Generator Type} & \textbf{Facility} \\ \hline
Zoom Range              & 0.99-1.01         \\ \hline
Brightness Range        & 0.8-1.2           \\ \hline
Fill Mode               & Constant          \\ \hline
Horizontal Flip         & True              \\ \hline
\end{tabular}
\end{table}

\subsubsection{Data Visualization: }
\begin{figure}[h]
    \centering
    \includegraphics[width=\linewidth]{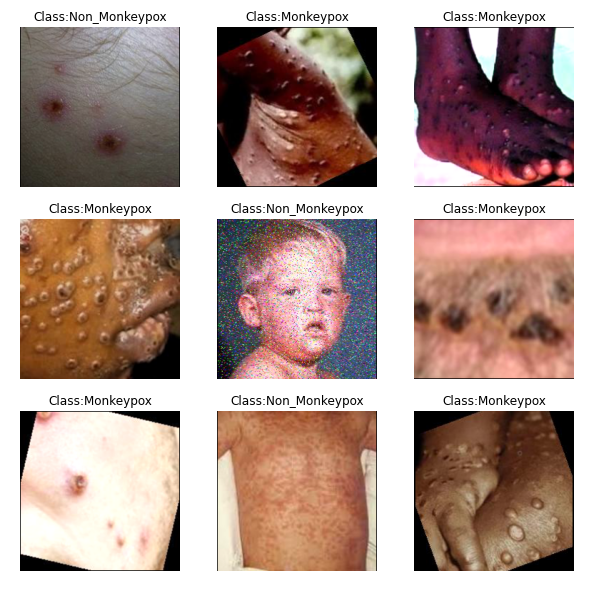}
    \caption{Random transformed image samples from the dataset.}
    \vspace{-10pt}
    \label{fig6}
\end{figure}

This stage requires sampling and reviewing the input data to observe the randomly transformed images and to verify that the transformations are being applied as expected. This stage is also used to confirm the readability of the input images. ImageDataGenerator library is used to generate a batch of randomly transformed images to visualize the transformed images. These transformed images are then visualized using the visualization library  Matplotlib via 2D representation with the new picture dimensions \cite{R27}. \textit{Figure \ref{fig6}} shows the visualization of the transformed image sample of our dataset.

\subsection{Splitting Dataset }

The dataset is split into two separate sets for the purpose of model training and testing, ensuring that the model is trained on a portion of the data and evaluated on a separate portion.
\begin{itemize}
\item Train data
\item Test data
\end{itemize}
Here, we have used 90\% (2872) of data for training the model and 10\% of data for testing the validation of the model. Traying several splitting ratios, we ended up with the splitting ratios 90\% and 10\% because model gives the better training accuracy and validation accuracy and also gives satisfactory results on other performance metrics for this splitting ratio.

\subsection{Model Building}

\begin{figure}[h]
    \centering
    \includegraphics[width=\linewidth]{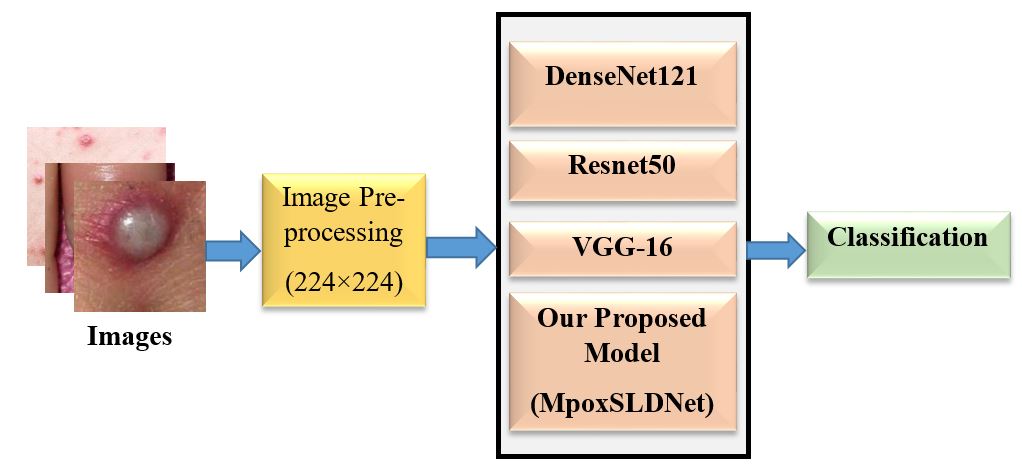}
    \caption{Pipeline to train the pre-trained model and MpoxSLDNet model.}
    \vspace{-10pt}
    \label{fig7}
\end{figure}

Monkeypox and non-monkeypox lesions can be classified using pre-trained models that were designed to identify 1000 classes in Imagenet. In this study, we have used three pre-trained including DenseNet121, ResNet50, and VGG16. Nevertheless, these pre-trained architectures have a lot of convolutional layers and take a lot of computation time and storage space. So, we proposed a lightweight convolutional network model named after 'MpoxSLDNet' to address these problems. This DL network 'MpoxSLDNet' uses a classification-based detection approach to extract relevant information from an input image, and improves it by adding layers (convolution, pooling, and dense). \textit{Table \ref{Table6}} shows the summary of models used in this study. The overall workflow to train the pre-trained model and MpoxSLDNet model are shown in \textit{Figure \ref{fig7}}

\begin{table*}[]
\caption{Summary of models used in this study.}
\label{Table4}
\centering
\begin{tabular}{|c|l|c|c|c|l|}
\hline
\textbf{\begin{tabular}[c]{@{}c@{}}Model \\ Name\end{tabular}} & \multicolumn{1}{c|}{\textbf{Description}}                                                                                                                                & \textbf{\begin{tabular}[c]{@{}c@{}}Batch \\ Size\end{tabular}} & \textbf{Epochs} & \textbf{Optimizer} & \multicolumn{1}{c|}{\textbf{\begin{tabular}[c]{@{}c@{}}Loss \\ Function\end{tabular}}} \\ \hline
MpoxSLDNet                                                     & \begin{tabular}[c]{@{}l@{}}A custom-built CNN model designed  for \\ the identification of Monkeypox and \\ Non-Monkeypox lesions.\end{tabular} & 32                                                             & 20              & Adam               & \begin{tabular}[c]{@{}l@{}}Binary \\ Cross entropy\end{tabular}                        \\ \hline
DenseNet121                                                    & \begin{tabular}[c]{@{}l@{}}A pre-trained CNN model with 121 layers \\ was used for feature extraction and \\ classification.\end{tabular}                                   & 32                                                             & 20              & Adam               & \begin{tabular}[c]{@{}l@{}}Binary \\ Cross entropy\end{tabular}                        \\ \hline
ResNet50                                                       & \begin{tabular}[c]{@{}l@{}}A pre-trained CNN model with 50 layers \\ is employed to extract features and perform \\ classification tasks.\end{tabular}                                     & 32                                                             & 20              & Adam               & \begin{tabular}[c]{@{}l@{}}Binary \\ Cross entropy\end{tabular}                        \\ \hline
VGG16                                                          & \begin{tabular}[c]{@{}l@{}}A pre-trained CNN model with 16 layers is \\ employed to extract features and perform \\ classification tasks.\end{tabular}                                     & 32                                                             & 20              & Adam               & \begin{tabular}[c]{@{}l@{}}Binary \\ Cross entropy\end{tabular}                        \\ \hline
\end{tabular}
\end{table*}

\subsubsection{Architecture of our Proposed MpoxSLDNet model :}
\begin{figure}[t]
    \centering
    \includegraphics[width=\linewidth]{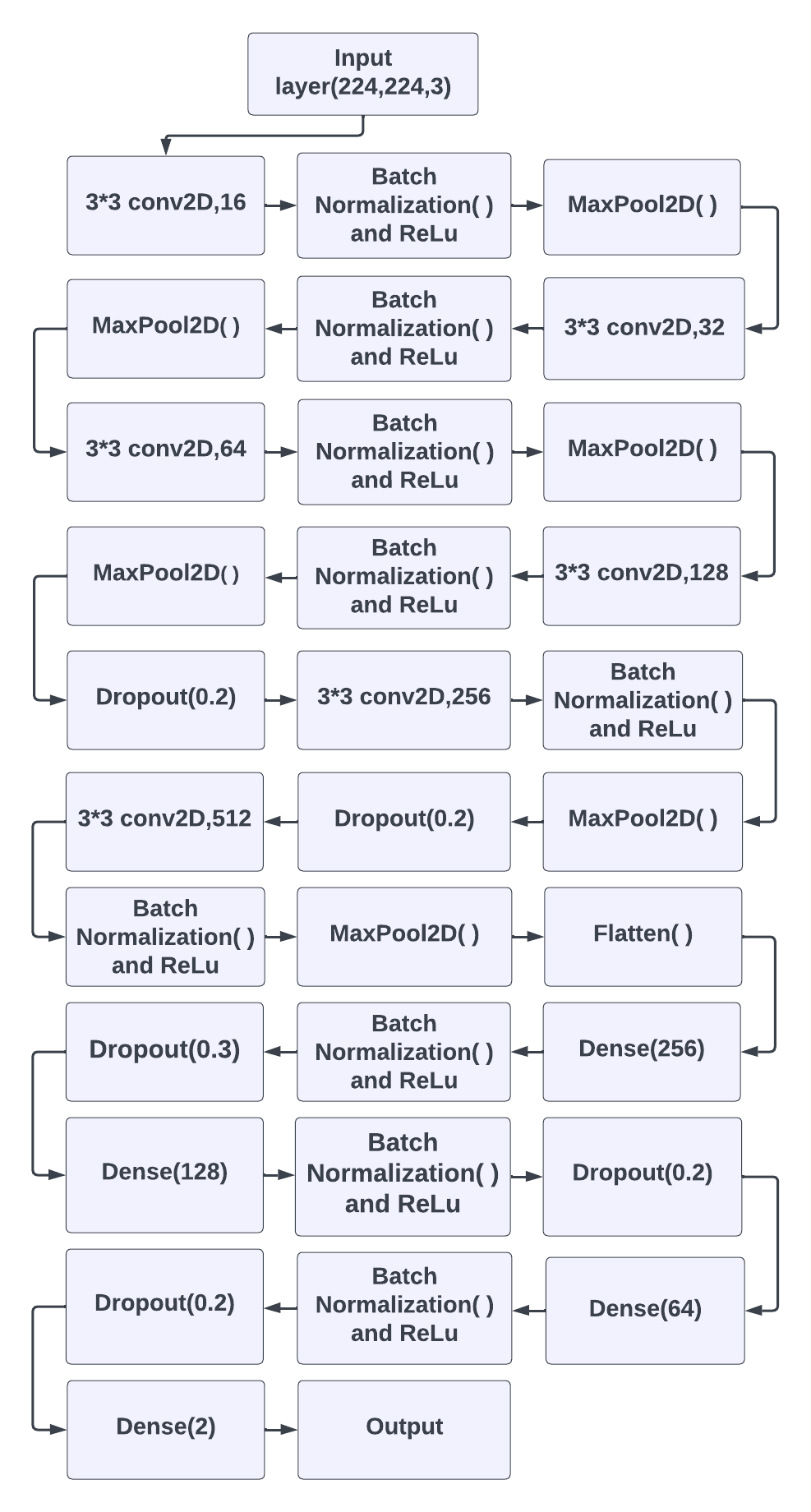}
    \caption{Architecture of proposed (MpoxSLDNet) model.}
    \vspace{-10pt}
    \label{fig8}
\end{figure}

\textit{Figure \ref{fig8}} shows the structure of the MpoxSLDNet model. This section provides a detailed description of the proposed (MpoxSLDNet) model's structure. 

\begin{figure}[h]
    \centering
    \includegraphics[width=\linewidth]{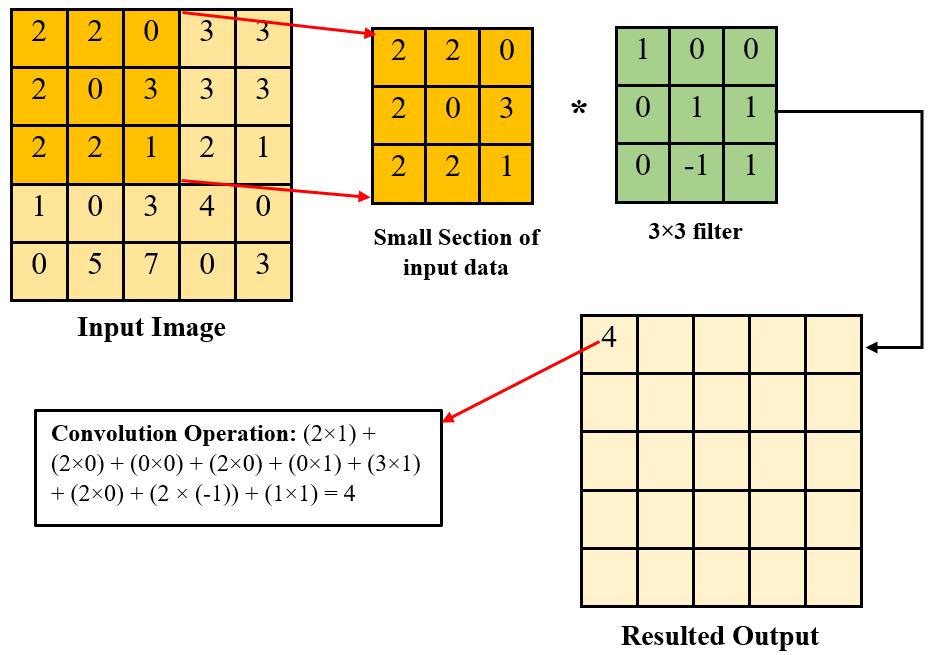}
    \caption{Example of the Convolution operation.}
    \vspace{-10pt}
    \label{fig9}
\end{figure}

\begin{figure}[h]
    \centering
    \includegraphics[width=0.8\linewidth]{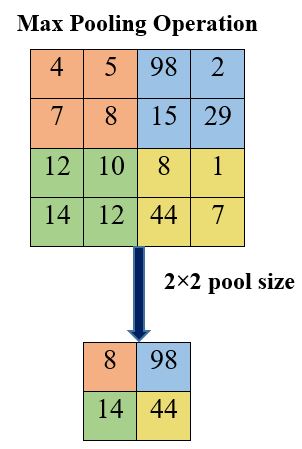}
    \caption{Illustration of Max Pooling.}
    \vspace{-10pt}
    \label{fig10}
\end{figure}

\begin{figure}[h]
    \centering
    \includegraphics[width=\linewidth]{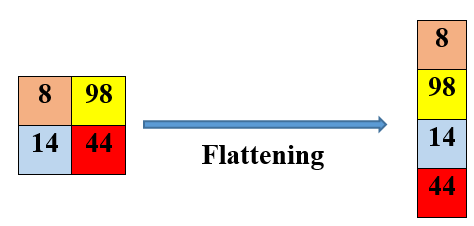}
    \caption{Converting 1D array.}
    \vspace{-10pt}
    \label{fig11}
\end{figure}

\begin{figure}[h]
    \centering
    \includegraphics[width=0.8\linewidth]{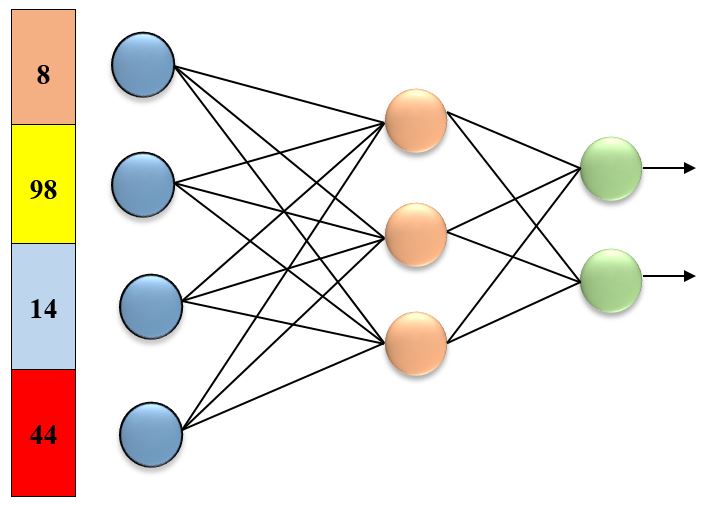}
    \caption{Fully Connected Layer.}
    \vspace{-10pt}
    \label{fig12}
\end{figure}

Our proposed model is a CNN architecture designed for 2-dimensional (2-D) image analysis that consists of six convolutional layers and six max-pooling layers. Input images with 224×224 dimensions are passed into the input layer. The initial convolutional layer consists of 16 feature kernel filters with a filter size of 3x3 and the padding is set to 'same'. The second convolutional layer consists of 32 feature kernel filters and filter size is 3×3 and padding=”same”. The third convolutional layer with a kernel size of 3×3 uses 64 feature maps. The fourth convolutional layer with a kernel size of 3×3 and use 128 feature maps. The fifth convolutional layer with a kernel size of 3×3 and use 256 feature maps. The next six convolutional layers with kernel size 3×3. These convolutional layers have 512 nodes. The resulting output dimensions of convolutional layers are 7x7x512. The ReLU activation function is applied after each of these convolutional layers to introduce non-linearity into the model which is important for learning complex features and patterns in the input data. The \textit{eqn \ref{eqn1}} for the ReLu activation function.
\begin{equation}
\label{eqn1}
   Y(x) = 
   \begin{cases}
   x,\hspace{0.5cm} if x\geq 0
   \\
   0,\hspace{0.5cm} Otherwise
   \end{cases}
\end{equation}
The convolutional layer performs the extraction of significant features from the input images, capturing valuable patterns and information. Here, The convolution operation can be mathematically expressed via \textit{eqn \ref{eqn2}}. 
\begin{equation}
\label{eqn2}
    {Y_{i,j}} = \sum_{s=1}^{S} \sum_{t=1}^T {W_{s,t}} {X_{(i+s-1),(j+t-1)}}
    \end{equation}
Output shape of the convolutional layer can be calculated by the \textit{eqn \ref{eqn3}}. 
\begin{equation}
\begin{split}
\label{eqn3}
    output\_shape = & ((input\_shape - filter\_size + 2 \\& * padding)
    / stride) +1
    \end{split}
\end{equation}
Consider a 5×5 image that convolved with a 3×3 filter where padding is zero; stride is one. The convolution operation is shown in \textit{Figure \ref{fig9}}. In the MpoxSLDNet model, After each conv2D layer, a BatchNormalization layer is utilized to normalize the output of the convolutional layer and to reduce overfitting. In this model, after each pair of conv2D and BatchNormalization layer MaxPool2D layer of a 2x2 filter and a stride of 2  is applied which downsamples the feature map by taking the maximum value within each pooling window, thereby reducing the dimensionality of the data. Here, the default parameters of padding for the MaxPool2D layer are valid. The max pooling operation can be represented mathematically by the \textit{eqn \ref{eqn4}}.
\begin{equation}
\label{eqn4}
    {Y_{i,j}} = max_{s=1}^{S} max_{t=1}^T {X_{(i+s-1),(j+t-1)}}
    \end{equation}
The shape of the output from the max pooling layer can be determined using \textit{eqn \ref{eqn5}}.
\begin{equation}
\begin{split}
\label{eqn5}
    output\_shape = & ((input\_shape - pool\_size )
    \\&/strides) +1
    \end{split}
\end{equation}
The \textit{Figure \ref{fig10}} provides an example of the max pooling operation using a 2x2 pixel filter on a 4x4 pixel input. After six convolution layers and six max-pooling layers, a flattened layer is added that flattens the output of the preceding layer into a one-dimensional array, facilitating its transfer to a fully connected layer because a fully associated layer accepts a 1D vector as input while convolution and pooling layer's output  is typically a three-dimensional tensor with dimensions representing the height, width, and number of channels. \textit{Figure \ref{fig11}} shows process of converting 1D array with the flatten layer. Then the converted 1D array is utilized by three fully connected layers of 256 units, 128 units, and 64 units, respectively, which are activated by the ReLu activation function. A Batch Normalization layer and dropout function are applied individually after each dense layer. A mathematical representation for a fully connected layer can be given as \textit{eqn \ref{eqn6}}.\begin{equation}
\label{eqn6}
    Y= f (Wx + b)
\end{equation}
\textit{Figure \ref{fig12}} shows the graphical representation of fully connected layer. Then the last layer is 2 units dense layer with a sigmoid activation function producing a binary classification output. \textit{Table \ref{Table4}} shows the summary of hyperparameters used in the proposed MpoxSLDNet model.

\begin{table}[]
\caption{Hyper parameters for the proposed MpoxSLDNet model.}
\centering
\label{Table5}
\begin{tabular}{|c|c|}
\hline
\textbf{Hyper Parameter}   & \textbf{Description} \\ \hline
No. Of Convolutional Layer & 6                    \\ \hline
No. Of Max pooling Layer   & 6                    \\ \hline
Activation Function        & ReLu, Sigmoid        \\ \hline
Batch Size                 & 32                   \\ \hline
Optimizer                  & Adam                 \\ \hline
Loss Function              & Binary Cross entropy \\ \hline
\end{tabular}
\end{table}

\section{Result Analysis}

In this study, the results obtained from training and testing all models on the dataset are presented. The performance of the models using various assessment measures, including confusion matrices, AUC-ROC score, Classification report, precision, recall, as well as F1-score is evaluated. Furthermore, a detailed discussion of the results and the strengths and weaknesses of each model in the context of monkeypox lesion detection are also provided. The original classifier of pre-trained architectures has been replaced. As a result of Kaggle's limited number of images, in-place data augmentation techniques were implemented during training to expand the dataset and solve the overfitting problem. After completing 20 training epochs, we recorded the accuracy and loss metrics for each model. 

\subsection{Assessing the MpoxSLDNet model's performance through multiple runs}

\textit{Table \ref{Table7}} shows the training and validation accuracy obtained from running the proposed model three times for 20 epochs. The purpose of multiple runs is to mitigate the effects of randomness during the training phase and provide a more reliable evaluation of how well the model performs. Based on the \textit{Table \ref{Table7}}, the MpoxSLDNet model attained an average training accuracy of 96.89\% and an average validation accuracy of 93.08\%. While the performance of the model is consistent across the three runs, there is some variation in the accuracy values. This highlights the importance of evaluating the model's performance multiple times to obtain a more accurate estimate of its accuracy.

\subsection{Accuracy and Loss Analysis }

The validation accuracies, training accuracies, and losses of each model are shown in \textit{Table \ref{Table8}}. Our proposed model (MpoxSLDNet) exhibited a training accuracy of 97.53\% and a training loss of 0.0862. The validation phase resulted in an accuracy of 94.56\% and a loss of 0.1583 for the MpoxSLDNet model. Similarly, the training phase for DenseNet121 resulted in an accuracy of 85.86\% and a loss of 0.3443. DenseNet121 achieved a validation accuracy of 84.38\% with a validation loss of 0.3280. Moving on to ResNet50, the training accuracy and loss were recorded as 72.42\% and 0.5333, respectively. The validation phase for ResNet50 resulted in an accuracy of 67.19\% and a loss of 0.5602. Similarly, The training accuracy for VGG16 was recorded as 89.45\%, with a corresponding training loss of 0.2404. VGG16 achieved a validation accuracy of 86.25\% with a validation loss of 0.3080. In summary, the MpoxSLDNet models that we proposed demonstrated superior performance in terms of both training and validation accuracy, outperforming other models such as DenseNet121, ResNet50, and VGG16.

\begin{table}[h]
    \caption{Multiple Training accuracy and Testing accuracy for the MpoxSLDNet model.}
    \resizebox{\linewidth}{!}{
\label{Table7}
\begin{scriptsize}
\centering
\begin{tabular}[c]{|  p{1.2cm} | p{0.8cm}  | p{.8cm}  | p{.8cm}  | p{.8cm}  | p{.8cm}  |}
\hline   

\hline
\textit{Experiment} & \textit{Epoch} & \textit{Training Accuracy} & \textit{Loss} & \textit{Testing 
Accuracy} & \textit{Loss}  \\ 
\hline

\hline
1  & 20 & 97.60\% &	0.0826 &93.12\% & 0.2199  \\
\hline
2  & 20 & 95.54\% &	0.1300 & 91.56\% & 0.2745 \\
\hline
3  & 20 & 97.53\% &	0.0862 & 94.56\% & 0.1583  \\
\hline

\end{tabular} 
\end{scriptsize} }
\end{table}

\begin{table}[h]
\caption{
Evaluation of accuracy and loss for each model during training and validation.}
    \resizebox{\linewidth}{!}{
\label{Table8}
\begin{tabular}[c]{|  p{4cm} | p{1.7cm}  | p{1.3cm}  | p{1.7cm}  | p{1.3cm}  |}

\hline
 \large Model Name                     & \large Training Accuracy & \large Loss   & \large Testing Accuracy & \large Loss   \\ \hline
\large DenseNet121                    & \large 85.86\%             & \large 0.3443 & \large 84.38\%          & \large 0.3280 \\ \hline
\large ResNet50                       & \large 72.42\%             & \large 0.5333 & \large 67.19\%          & \large 0.5602 \\ \hline
\large VGG16                          & \large 89.45\%             & \large 0.2404 & \large 86.25\%          & \large 0.3080 \\ \hline
\large Our proposed Model (MpoxSLDNet) & \large 97.53\%             & \large 0.0862 & \large 94.56\%          & \large 0.1583 \\ \hline
\end{tabular}}
\end{table}

\begin{figure}[h]
\centering
    \includegraphics[width=\linewidth]{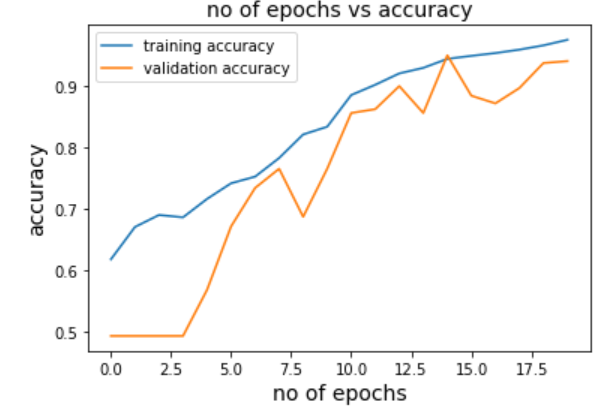}
    \caption{Representation of accuracy curve of our proposed (MpoxSLDNet) model.}
    \vspace{-10pt}
    \label{fig13}
\end{figure}

\begin{figure}[h]
\centering
    \includegraphics[width=\linewidth]{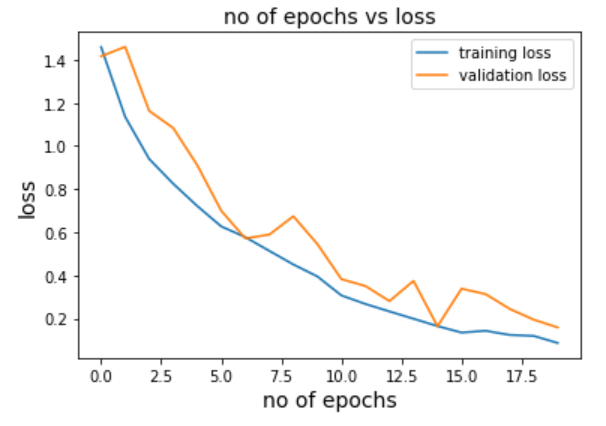}
    \caption{Representation of loss curve of our proposed (MpoxSLDNet) model.}
    \vspace{-10pt}
    \label{fig14}
\end{figure}

\textit{Figure \ref{fig13}} as well as \textit{Figure \ref{fig14}} shows the accuracy and loss learning curve of our proposed model (MpoxSLDNet).
The learning curve illustrates the evolution of the model's performance changes with training. Learning curves provide valuable insights into how the model behaves and help us fine-tune the hyperparameters for optimal performance, as well as identify the model's generalization capability, indicating how well it adapts to unseen data or performs well. From the accuracy as well as loss curve, we can observe that the validation and accuracy curves start to converge as the training goes on, showing that the model is learning and improving its performance and whether it is overfitting or underfitting the data. By analyzing the learning curve, we can make informed decisions that the model is starting to generalize well on the unseen data, which is a good sign of model performance. 

\begin{table}[h]
    \caption{Comparative Classification performance analysis of our proposed model vs. pre-trained model.}
    \resizebox{\linewidth}{!}{
\label{Table9}
\begin{scriptsize}
\centering
\begin{tabular}[c]{|  p{2.2cm} | p{1.8cm}  | p{1cm}  | p{.8cm}  | p{.8cm}  |}
\hline   

\hline
\textit{Model Name} & \textit{Patient Status} & \textit{Precision} & \textit{Recall} & \textit{F1-score}  \\ 
\hline

\hline
\multirow{2}{*}{DenseNet121}  & Monkeypox(0) & 0.88 &	0.80 & 0.84\\ \cline{2-5}
 & Non-Monkeypox(1) & 0.81 &	0.89 & 0.85\\ \cline{2-5}
\hline
\multirow{2}{*}{ResNet50}  & Monkeypox(0) & 0.86 &	0.42 & 0.56\\ \cline{2-5}
 & Non-Monkeypox(1) & 0.61 &	0.93 & 0.74\\ \cline{2-5}
 \hline
\multirow{2}{*}{VGG16}  & Monkeypox(0) & 0.94 &	0.78 & 0.85\\ \cline{2-5}
 & Non-Monkeypox(1) & 0.81 &	0.95 & 0.87\\ \cline{2-5}
 \hline
\multirow{2}{*}{MpoxSLDNet model}  & Monkeypox(0) & 0.94 &	0.94 & 0.94\\ \cline{2-5}
 & Non-Monkeypox(1) & 0.94 &	0.94 & 0.94\\ \cline{2-5}
 \hline
\end{tabular} 
\end{scriptsize} }
\end{table}

\subsection{Classification Report Analysis  }

The classification report gives a detailed breakdown of precision, recall, and F1 scores of each class. Using the classification report, we looked at precision, recall, and F1-score for each class to assess the proposed MpoxSLDNet model's performance. \textit{Table \ref{Table9}} presents the classification report for each models used in this study, including DenseNet121, ResNet50, VGG16, and our proposed model (MpoxSLDNet). The classification report includes precision, recall, and F1-score values for each class, namely Monkeypox (0) and Non-Monkeypox (1), indicating the model's performance in detecting monkeypox lesions. For DenseNet121, the precision for Monkeypox is 0.88, indicating that 88\% of the samples classified as Monkeypox were correctly identified. The recall for Monkeypox is 0.80, indicating that 80\% of the actual Monkeypox samples were correctly identified. The F1-score for Monkeypox is 0.84, providing a balanced measure of precision and recall. Similarly, for Non-Monkeypox class, the precision, recall, and F1-score values are provided. The same evaluation metrics are provided for ResNet50 and VGG16 models, highlighting their performance in detecting Monkeypox and Non-Monkeypox lesions. Our proposed model, MpoxSLDNet, demonstrates superior performance in both classes, with high precision, recall, and F1-score values of 0.94 for both the Monkeypox and Non-Monkeypox classes. This indicates that the MpoxSLDNet model achieved a high level of accuracy in correctly identifying both types of lesions. This classification report table allows for a comprehensive comparison of the models' performance in detecting monkeypox lesions, providing insights into the precision, recall, and F1-score values for each class. These metrics help to evaluate and select the most effective model for accurate monkeypox lesion detection.

\begin{figure}[h]
\centering
    \includegraphics[width=\linewidth]{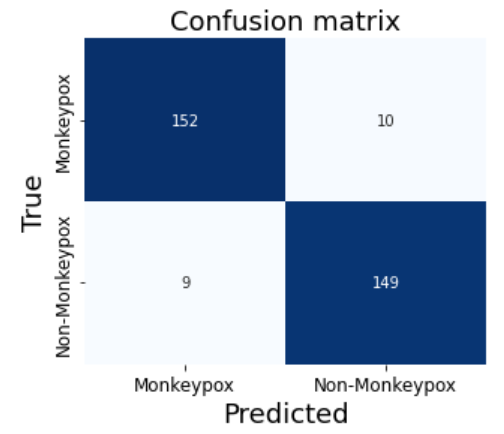}
    \caption{Confusion Matrix for our proposed (MpoxSLDNet) model.}
    \vspace{-10pt}
    \label{fig15}
\end{figure}

\begin{figure}[h]
\centering
    \includegraphics[width=\linewidth]{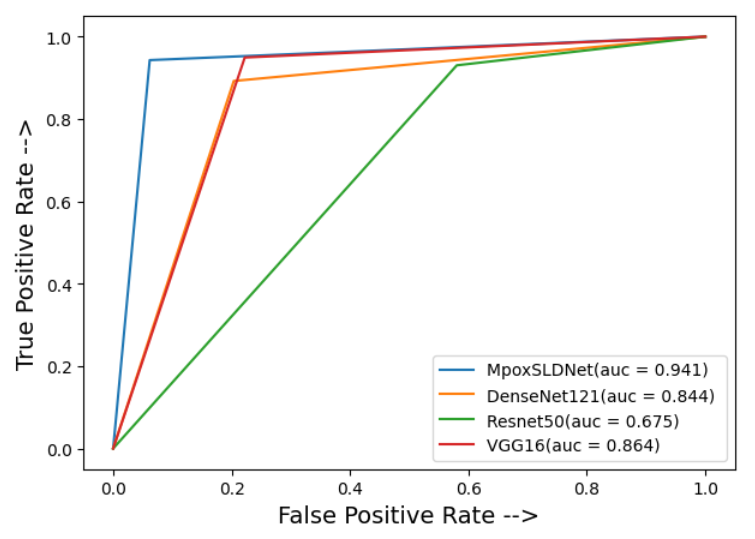}
    \caption{AUC-ROC curve for all models.}
    \vspace{-10pt}
    \label{fig16}
\end{figure}

\subsection{Confusion Matrix and AUC ROC Curve Analysis }
Confusion matrices are good tools for evaluating classification models. This confusion matrix shows how many correctly classified and misclassified samples there are for each class. \textit{Figure \ref{fig15}} shows the confusion matrix of our proposed MpoxSLDNet model. Here, 320 of the test images are evaluated. True Positive tells us how often the model correctly categorizes monkeypox. In the same way, True Negative denotes the model's ability to correctly classify non-monkeypox as non-monkeypox. From \textit{Figure \ref{fig15}}, we can observe that the MpoxSLDNet model accurately identifies 152 images as monkeypox (True positive) and 149 images as non-monkeypox (True Negative). According to {Figure \ref{fig15}}, we can conclude that our model recognized 301 images correctly out of 320. On the other hand, false positive means that the model incorrectly classifies non-monkeypox as monkeypox. Conversely, False negative shows  the instances where the model mistakenly classifies  monkeypox as non-monkeypox. We can also observe that 9 images of Non-Monkeypox were incorrectly classified as monkeypox positive(False Positive), and 10 images of Monkeypox were incorrectly classified as Non-Monkeypox(negative). It shows that our approach can accurately identify Monkeypox patients, with high true positives and true negatives. Additionally, we employed Receiver Operating Characteristic (ROC) curves to evaluate the classification performance which represents the true positive rate (TPR) plotted against the false positive rate (FPR), aiding in the detection of positive Monkeypox cases within the tested Monkeypox and Non-Monkeypox images. Here, AUC measures how much our model is capable of distinguishing between our two classes, Monkeypox and Non-Monkeypox. The AUC-ROC curve is displayed for all the employed models in \textit{Figure \ref{fig16}}. According to the data provided in \textit{Table \ref{Table10}}, the proposed model demonstrated an AUC-ROC score of 0.94. The AUC-ROC score of the DenseNet121, ResNet50, and VGG16 are 0.84, 0.68, and 0.86, respectively. It is observed that the best AUC-ROC score belongs to the MpoxSLDNet model. Hence, the MpoxSLDNet model is relatively effective at distinguishing between positive and negative instances. 

\subsection{Overall Performance Evaluation and Comparative Analysis: Proposed Model vs. Pre-trained Models }

\begin{figure}[h]
\centering
    \includegraphics[width=\linewidth]{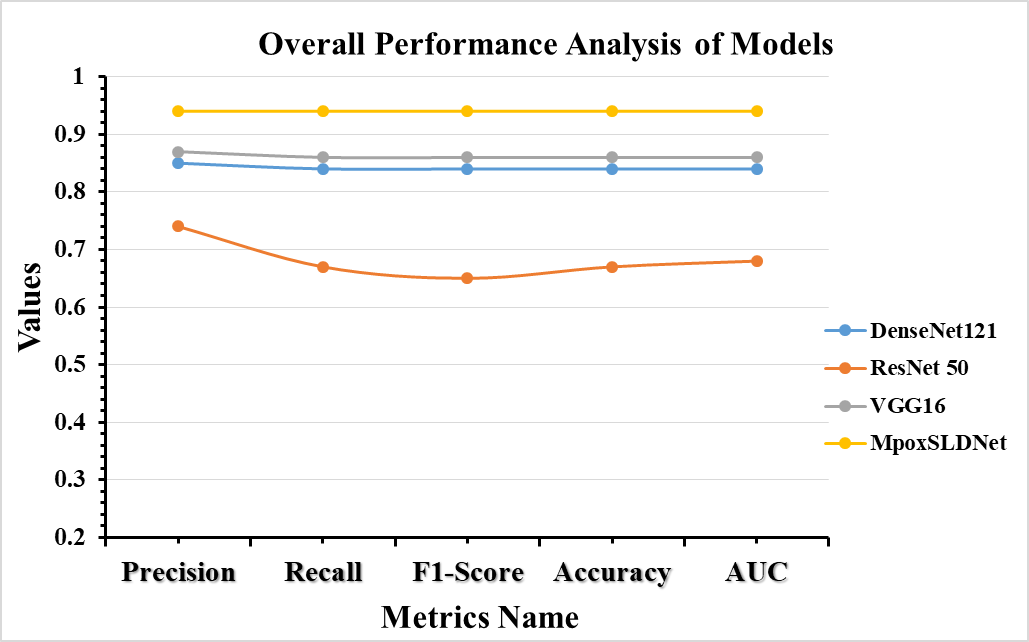}
    \caption{Overall performance Comparison of MpoxSLDNet model vs. pre-trained model.}
    \vspace{-10pt}
    \label{fig17}
\end{figure}

In this section, we present an overall comprehensive analysis and comparison of the performance between our proposed model, MpoxSLDNet, and pre-trained models. The evaluation is conducted using key metrics including accuracy, precision, recall, and F1-score, AUC score. As shown in \textit{Table \ref{Table10}}, the highest precision score of 0.94 occupied by the MpoxSLDNet model indicates that the model is good at correctly identifying positive instances. On the other hand, as recall increases, monkeypox images are identified more frequently, and the opposite is also true. With a recall score of 0.94, the MpoxSLDNet model reflects its ability to correctly identify a significant number of monkeypox images. Apart from this, the F1 score combines precision and recall into one score. The F1-score achieves its highest value of 1, indicating excellent performance, and its lowest value of 0, indicating poor performance. The MpoxSLDNet model achieves the highest average F1 score of 0.94, indicating a superior balance between precision and recall compared to the pre-trained model. Furthermore, the accuracy metric measures how many instances are classified correctly. With an accuracy score of 0.94, the model was able to classify instances correctly. 
\textit{Figure \ref{fig17}} serves as a succinct overview of the model's performance, allowing for easy comparison and identification of the top-performing model according to the selected metrics. The x-axis represents the metrics name as well as y-axis represents the corresponding values for precision, recall, F1-score, accuracy, and AUC score ranging from 0 to 1. The graph provides a visual representation of how each model performed in terms of these metrics. Higher values indicate better performance. Undoubtedly, MpoxSLDNet consistently demonstrates superior precision, recall, F1-score, and accuracy compared to the other models. This suggests that the MpoxSLDNet model surpasses the other models in accurately detecting monkeypox lesions.

\subsection{Evaluating the Performance of the Proposed Model in Comparison to Existing Research }

\textit{Table \ref{Table11}}  presents a comparative analysis between the existing study and the proposed system, both utilizing the Monkeypox Skin Lesion Dataset (MSLD). The accuracy scores, represented as percentages, reflect the models' ability to accurately classify monkeypox lesions from non-monkeypox lesions. Our proposed MpoxSLDNet model demonstrates an accuracy score of 94.56\%, surpassing all other models in accuracy. In comparison, the previously employed models exhibited accuracy scores ranging from 71.86\% to 91.11\%. Apart from, in this study, other pre-trained models like DenseNet121, ResNet50, and VGG16 demonstrate accuracy values ranging from 67.19\% to 86.25\%. This also highlights the superior accuracy achieved by MpoxSLDNet compared to the other models, showcasing its effectiveness in accurately classifying monkeypox lesions. \textit{Table \ref{Table11}} effectively summarizes the accuracy performance of the different models, highlighting the superior performance of MpoxSLDNet in accurately detecting and classifying monkeypox lesions.

\begin{table}[h]
    \caption{Overall performance of our proposed model vs. pre-trained model.}
    \resizebox{\linewidth}{!}{
\label{Table10}
\begin{scriptsize}
\centering
\begin{tabular}[c]{|  p{2.3cm} | p{1.3cm}  | p{.8cm}  | p{1.3cm}  | p{1cm}  | p{.8cm}  |}
\hline   

\hline
\textit{Model Name} & \textit{Precision} & \textit{Recall} & \textit{F1-score} & \textit{Accuracy} & \textit{AUC-ROC Value}  \\ 
\hline

\hline
DenseNet121  & 0.85 & 0.84 & 0.84 & 0.84 & 0.84  \\
\hline
ResNet50  & 0.74 & 0.67 & 0.65 & 0.67 & 0.68  \\
\hline
VGG16  & 0.87 & 0.86 &	0.86 & 0.86 & 0.86  \\
\hline
Our proposed Model (MpoxSLDNet)  & 0.94 & 0.94 &	0.94 & 0.94 & 0.94  \\
\hline

\end{tabular} 
\end{scriptsize} }
\end{table}

\begin{table}[]
 \caption{Result Analysis of existing study with the proposed system.}
 \label{Table11}
 \centering
\begin{tabular}{|c|c|}
\hline
\textbf{Model Name}                                    & \textbf{Accuracy} \\ \hline
MobileNetv2  \hspace{.03 cm} \cite{R1}                & 91.11\%           \\ \hline
VGG16 \hspace{.03 cm}  \cite{R11}                     & 81.48\%           \\ \hline
ResNet50 \hspace{.03 cm}  \cite{R11}                  & 74.07\%           \\ \hline
InceptionV3 \hspace{.03 cm}  \cite{R11}               & 82.96\%           \\ \hline
Ensemble \hspace{.03 cm}  \cite{R11}                  & 79.26\%           \\ \hline
VGG19-CBAM-Dense \hspace{.03 cm}  \cite{R21}          & 71.86\%           \\ \hline
DenseNet121-CBAM-Dense \hspace{.03 cm}  \cite{R21}    & 78.27\%           \\ \hline
EfficientNetB3-CBAM-Dense\hspace{.03 cm}  \cite{R21} & 81.43\%           \\ \hline
MobileNetV2-CBAM-Dense\hspace{.03 cm}  \cite{R21}    & 74.07\%           \\ \hline
Xception-CBAM-Dense \hspace{.03 cm} \cite{R21}       & 83.89\%           \\ \hline
Proposed Model (MpoxSLDNet)                            & 94.56\%           \\ \hline
VGG16                                                  & 86.25\%           \\ \hline
DenseNet121                                            & 84.38\%           \\ \hline
ResNet50                                                 & 67.19\%           \\ \hline
\end{tabular}
\end{table}

\section{Discussion}

The study's outcomes highlight the substantial progress made in monkeypox lesion detection with the development and evaluation of the MpoxSLDNet model. By focusing on key performance metrics like accuracy, precision, recall, and F1-score, our results consistently demonstrate the MpoxSLDNet model's superiority over traditional pre-trained architectures such as DenseNet121, ResNet50, and VGG16. Through rigorous validation and multiple runs, our model consistently achieves high training and validation accuracies, indicative of its robustness and reliability in discerning monkeypox lesions from digitized skin lesion images. Additionally, the use of confusion matrices and AUC-ROC curves further validates the model's efficacy in distinguishing between positive and negative instances, with the MpoxSLDNet model consistently outperforming others, as evidenced by its highest AUC-ROC score. These findings hold significant implications for disease management, providing healthcare practitioners with a powerful tool for early detection and intervention in monkeypox cases, thereby enabling prompt treatment and containment measures to curb disease transmission. So we can say, MpoxSLDNet model represents a groundbreaking advancement in the realm of monkeypox lesion detection, offering versatile applications with profound potential impact. Primarily, it enables early detection and diagnosis of monkeypox lesions, aiding in timely intervention and treatment to mitigate disease progression and transmission. Its non-invasive nature makes it particularly advantageous in healthcare settings, minimizing physical contact and reducing the risk of disease spread among healthcare professionals and patients. By providing automated lesion detection and classification, the model enhances clinical decision-making, improving diagnostic accuracy and treatment outcomes. Furthermore, its integration into public health surveillance systems holds promise for early outbreak detection, epidemiological monitoring, and targeted intervention strategies. Beyond healthcare, the MpoxSLDNet model may influence industries involved in disease surveillance, pharmaceutical research, and public health policy, informing efforts towards developing treatments, vaccines, and policy decisions for disease control and prevention. Overall, the MpoxSLDNet model's deployment has the potential to significantly enhance healthcare delivery, disease control efforts, and public health outcomes on a global scale.

\section{Conclusion}

In conclusion, this study introduced a deep learning-based framework for effectively classifying skin lesions associated with monkeypox and non-monkeypox that achieved promising results. Through comprehensive experimentation and comparison with established pre-trained models, including VGG16, ResNet50, and DenseNet121, our research aimed to evaluate MpoxSLDNet's effectiveness. The findings demonstrate MpoxSLDNet's superior performance metrics, including accuracy, precision, recall, F1-score, and AUC, underscoring its potential as a reliable tool for early detection and categorization of monkeypox lesions. \textit{Table \ref{Table9}} demonstrates that the suggested MpoxSLDNet model showed superior performance compared to traditional pre-trained models, with higher accuracy, precision, recall, F1-score, and AUC metrics. The proposed MpoxSLDNet model achieved remarkable results, including an accuracy rate of 94.56\%, a precision score of 0.94, a recall rate of 0.94, an F1-score of 0.94, and an AUC value of 0.94. In comparison, VGG16, DenseNet121, and ResNet50 obtained accuracy scores of 86.25\%, 84.38\%, and 67.19\% respectively.Despite these promising outcomes, it is crucial to acknowledge inherent limitations in our study, particularly concerning the dataset's representativeness of real-world clinical scenarios. Future research efforts should prioritize expanding the dataset to include a more diverse range of lesion variations and conducting rigorous clinical validation studies.

\section{Limitation and Future Work}

Despite the promising outcomes achieved with the Mpox- SLDNet model, it is important to acknowledge certain limitations inherent in our study. The dataset used for training and evaluation, while carefully curated, may not fully represent the diverse range of lesion variations encountered in real-world clinical settings. This limitation could potentially impact the model's ability to generalize to unseen cases, particularly those with rare or atypical lesion characteristics.
To address these limitations and further enhance the MpoxSLDNet model's effectiveness, expanding the dataset to include a more diverse range of lesion types, sizes, colors, and textures is imperative to improve the model's robustness and generalizability across different patient populations and clinical contexts. Additionally, conducting rigorous clinical validation studies to assess the model's performance in real-world settings is crucial.

\bibliographystyle{IEEEtran}

\appendices

\end{document}